\documentclass{article}
\usepackage{spconf,amsmath,graphicx}

\title{MULTI-MODAL EMOTION RECOGNITION ON IEMOCAP WITH NEURAL NETWORKS.}
\begin{document}

\name{Samarth Tripathi$^{\ddagger}$ \qquad Sarthak Tripathi$^{\star}$ \qquad Homayoon Beigi$^{\dagger}$}

\address{$^{\ddagger}$ Advanced AI, LG Silicon Valley Lab\\
	Santa Clara, California, USA-94041\\
	{\tt samarthtripathi@gmail.com} \\
	$^{\star}$ Uber Technologies, Bangalore, India\\
	{\tt sarthaktripathi3.gmail.com}\\
	$^{\dagger}$ Recognition Technologies, Inc.\\
	South Salem, New York, USA-10590  \\
	{\tt beigi@recotechnologies.com } \\
	}
%

%
\maketitle
\begin{abstract}
Emotion recognition has become an important field of research in human computer interactions and there is a growing need for automatic emotion recognition systems. One of the directions the research is heading is the use of neural networks which are adept at estimating complex functions that depend on a large number and diverse source of input data. In this paper we attempt to exploit this effectiveness of neural networks to enable us to perform multimodal emotion recognition on IEMOCAP dataset using data from speech, text, and motions captured from face expressions, rotation and hand movements. Our approach first identifies best individual architectures for classification on each modality and performs fusion only at the final layer which allows for a more robust and accurate emotion detection.
\end{abstract}
\section{Introduction}
\label{sec:intro}

Emotion is a psycho-physiological process that can be triggered by conscious and/or unconscious perception of objects and situations, associated with multitude of factors such as mood, temperament, personality, disposition, and motivation \cite{Soleymani}. Emotions are very important in human decision handling, interaction and cognitive process \cite{Sreeshakthy}. With the advancement of technology and as our understanding of emotions is advancing, there is a growing need for automatic emotion recognition systems. Emotion recognition has been studied widely using speech \cite{Lee} \cite{Chernykh} \cite{Neumann}, text \cite{Kim}, facial cues \cite{Bassili}, and EEG based brain waves \cite{Tripathi} individually.  One of the biggest open-sourced multimodal resources available in emotion detection is IEMOCAP dataset \cite{Busso} which consists of approximately 12 hours of audio-visual data, including facial recordings, speech and text transcriptions. 

In this paper we combine these modes to make a stronger and more robust detector for emotions. We explore various deep learning based architectures to first get the best individual detection accuracy from each of the different modes. We then combine them in an ensemble based architecture to allow for training across the different modalities using the variations of the better individual models. Our ensemble consists of Long Short Term Memory networks, Convolution Neural Networks, fully connected Multi-Layer Perceptrons and we complement them using techniques such as Dropout, adaptive optimizers such as Adam, pretrained word-embedding models and Attention based RNN decoders. This allows us to individually target each modality and only perform feature fusion at the final stage. The advantages of our study are two-fold. First, since we target each modality individually, lack of availability of any modality does not cripple our algorithm and would not require retraining of other modalities but only the prefinal layer. This also allows our approach to be modular. Second, we use Motion-capture data instead of Video recording, hence we do not use 3D-Convolutions but 2D-Convolutions which are faster have less memory requirements. We also use advanced hyperparameter optimization tools to achieve the best possible model configuration depending on our resource constraints. Our code is open sourced for other researchers to repeat and enhance our study.

\section{Related Works}
\label{sec:prior}

Most of the early research on IEMOCAP has concentrated specifically on emotion detection using speech data. One of the early important papers on this dataset is \cite{Han} where they used segment level feature extraction, to feed those features to a MLP based architecture, where the input is 750 dimensional feature vector, followed by 3 hidden layer of 256 neurons each with rectilinear units as non-linearity. \cite{Lee} follows \cite{Han} and they train long short-term memory (LSTM) based recurrent neural network. First they divide each utterance into small segments with voiced region, then assume that the label sequences of each segment follows a Markov chain. They extract 32 features for every frame with 12-dimensional Mel-frequency cepstral coefficients (MFCC) with log energy, and their first time derivatives among others. The network contains 2 hidden layers with 128 BLSTM cells (64 forward nodes and 64 backward nodes).

Another research we closely follow is \cite{Chernykh}, where they use CTC loss function to improve upon RNN based Emotion prediction. They use 34 features including 12 MFCC, chromagram-based and spectrum properties like flux and roll-off. For all speech intervals they calculate features in 0.2 second window and moving it with 0.1 second step. The use of CTC loss helps, as often, almost the whole utterance has no emotion, but emotionality is contained only in a few words or phonemes in an utterance which the CTC loss handles well. Unlike \cite{Lee} which uses only the improv data, Chernykh et. al. use all the session data for the emotion classification. 

Multi-modal emotion classification has recently gathered more traction and IEMOCAP remains the significant dataset for this research direction. The current state-of-art classification on IEMOCAP is provided by \cite{poria2018multimodal} which builds on the prior work \cite{poria2017context}. They use  3D-CNN for visual feature extraction, text-CNN for textual features extraction and openSMILE for audio feature extraction. They use Contextual LSTM Architecture on top of these unimodal input features. They are then merged with multi-modal contextual LSTM layers which performs feature fusion. This layer finally feeds to the classification module. In our paper we adopt a different approach to this study and achieve similar performance with certain advantages.   

\section{Experimental Setup}
\label{sec:RW}

IEMOCAP has 12 hours of audio-visual data from 10 actors where the recordings follow dialogues between a male and a female actor in both scripted or improvised topics. After the audio-visual data has been collected it is divided into small utterances of length between 3 to 15 seconds which are then labelled by evaluators. Each utterance is evaluated by 3-4 assessors. The evaluation form contained 10 options (neutral, happiness, sadness, anger, surprise, fear, disgust frustration, excited, other). We consider only 4 of them — anger, excitement (happiness), neutral and sadness so as to remain consistent with the prior research. We consider emotions where atleast 2 experts were consistent with their decision, which is more than 70 \% of the dataset, consistent with prior research.

Along with the .wav file for the dialogue we also have the transcript for each the utterance. For each session, one actor wears the Motion Capture (MoCap) camera data which records the facial expression, head and hand movements of the actor. The Mocap data contains column tuples, for facial expressions the tuples are contained in 165 dimensions, 18 for hand positions and 6 for head rotations. As this Mocap data is very extensive we use it instead of the video recording in the dataset. These three modes (Speech, Text, Mocap) of data form the basis of our multi-modal emotion detection pipeline. 

Next we preprocess the IEMOCAP data for these modes. For the speech data our preprocessing follows the work of \cite{Chernykh}. We use the Fourier frequencies and energy-based features Mel-frequency Cepstral Coefficients (MFCC) for a total of 34 features. They include 13 MFCC, 13 chromagram-based and 8 Time Spectral Features like zero crossing rate, short-term energy, short-term entropy of energy, spectral centroid and spread, spectral entropy, spectral flux, spectral rolloff. We calculate features in 0.2 second window and moving it with 0.1 second step and with 16 kHz sample rate. We keep a maximum of 100 frames or approximately for 10 seconds of the input, and zero pad the extra signal and end up with (100,34) feature vector for each utterance. We also experiment with delta and double-delta features of MFCC but they dont produce any performance improvement while adding extra computation overhead.

For the text transcript of each of the utterance we use pretrained Glove embeddings \cite{Penn} of dimension 300, along with the maximum sequence length of 500 to obtain a (500,300) vector for each utterance. For the Mocap data, for each different mode such as face, hand, head rotation we sample all the feature values between the start and finish time values and split them into 200 partitioned arrays. We then average each of the 200 arrays along the columns (165 for faces, 18 for hands, and 6 for rotation), and finally concatenate all of them to obtain (200,189) dimension vector for each utterance. 

The total dataset consists of 4936 dialogues. For individual modalities we divide our dataset with a randomly chosen 20\% validation splits. For the final combined model we use 3838 (77.7\%) as our training set, these correspond to first 4 sessions of the data with 8 actors. We use the final 1098 (22.2\%) dialogues as our test set. These correspond to Session 5, with 2 actors (Male and Female). This ensures we remain speaker agnostic in our predictions. Unlike \cite{poria2018multimodal} we do not use 10-fold cross validation, since cross validation on Neural Networks is unfeasible due to time and compute requirements. For HyperParameter Optimization (HPO) we use Auptimizer  \footnote{https://github.com/LGE-ARC-AdvancedAI/auptimizer} an open-sourced HPO tool. We use 838 dialogues from 3838 training set as our validation set for HPO. Once best parameters have been found we use both training and validation to evaluate on the test set.

\section{Models}
\label{sec:models}

\subsection{Speech Based Emotion Detection}

Our first model - Speech\_Model1 consists of three layered fully connected MLP layers with 1024, 512, 256 hidden neural units with Relu as activation and 4 output neurons with Softmax (like \cite{Han}). The model takes the flattened speech vectors as input and trains using cross entropy loss with Adadelta as the optimizer. Speech\_Model2 uses two stacked LSTM layers with 512 and 256 units followed by a Dense layer with 512 units and Relu activation (like \cite{Lee}). Speech\_Model3 uses 2 LSTM layers with 128 units each but the second LSTM layer has Attention implementation as well, followed by 512 units of Dense layer with Relu activation. Speech\_Model4 improves both the encoding LSTM and Attention based decoding LSTM by making them bi-directional. All the last 3 models use Adadelta as the optimizer. As we can see the final Attention based bidirectional LSTM model performs the best. We also try many variations of the speech data including using MelSpectrogram, smaller window (0.08s) with longer context (200 timestamps) but do not achieve improvements. 

\begin{table}[!t]
\renewcommand{\arraystretch}{1.3}
\caption{Speech emotion detection models and accuracy}
\label{table_spch}
\centering
\begin{tabular}{|c||c|}
\hline
Model & Accuracy\\
\hline
Speech\_Model1 & 50.6\% \\
Speech\_Model2 & 51.32\%\\
Speech\_Model3 & 54.15\%\\
Speech\_Model4 & 55.65\%\\
\hline
\end{tabular}
\end{table}

\begin{table}[!t]
\renewcommand{\arraystretch}{1.3}
\caption{Comparison between our Speech emotion detection models and previous research}
\label{table_spch_com}
\centering
\begin{tabular}{|c||c|}
\hline
Model & Accuracy\\
\hline
Lee and Tashev \cite{Lee} & 62.85\% \\
Ours (improv only) & 62.72\%\\
\hline
Chernykh \cite{Chernykh} & 54\% \\
Neumann \cite{Neumann} & 56.10\% \\
Lakomkin \cite{Lakomkin} & 56\% \\
Ours (all) & 55.65\% \\
\hline
\end{tabular}
\end{table}

To compare our results with prior research we use our best model (Speech\_Model4) and evaluate it in the manner similar to various conditions of the previous researches. Like \cite{Lee} we use only the improvisation session for both training and testing and achieve similar results. To compare with \cite{Chernykh} \cite{Neumann} \cite{Lakomkin} who use the both scripted and improvisation sessions we again achieve similar results. One important insight of our results is with minimal preprocessing and no complex loss functions or noise injection into the training, we can easily match prior research's performance using Attention based Bidirectional LSTMs. 

\subsection{Text based Emotion Recognition}
Our task of performing emotion detection using only the text transcripts of our data resembles that of sentiment analysis, a very common and well researched task of Natural Language Processing. Here we try two approaches Text\_Model1 which uses 1D convolutions of kernel size 3 each, with 256, 128, 64 and 32 filters using Relu as Activation and Dropout of 0.2 probability, followed by 256 dimension fully connected layer and Relu, feeding to 4 output neurons with Softmax. Text\_Model2 uses two stacked LSTM layers with 512 and 256 units followed by a Dense layer with 512 units and Relu Activation. Both these models are initialized with Glove Embeddings based word-vectors. We also try Randomized initialization with 128 dimensions in Text\_Model3 and obtain similar performance as Text\_Model2. The LSTM based models use Adadelta and Convolution based models use Adam as optimizers.

\begin{table}[!t]
\renewcommand{\arraystretch}{1.3}
\caption{Text emotion detection models and accuracy}
\label{table_text}
\centering
\begin{tabular}{|c||c|}
\hline
Model & Accuracy\\
\hline
Text\_Model1 & 62.55\% \\
Text\_Model2 & 64.68\%\\
Text\_Model3 & 64.78\%\\
\hline
\end{tabular}
\end{table}

\subsection{MoCap based Emotion Detection}
For the Mocap based emotion detection we use LSTM and Convolution based models. For emotion detection using only the head rotation we try 2 models, Head\_Model1 uses LSTM with 256 units followed by Dense layer and Relu activation, while Head\_Model2 uses just 256 hidden unit based Dense Layer with Relu and achieves better performance. We use the two models again for Hand movement based emotion detection and Hand\_Model2 again achieves better performance. For the facial expression based Mocap data (which has a larger dimensionality than Mocap head and hand data), we use two stacked LSTM layers with 512 and 256 units followed by a Dense layer with 512 units and Relu Activation as Face\_Model1. Face\_Model2 uses 5 2D Convolutions each with kernel size 3, Stride 2 and 32, 64, 64, 128, 128 filters, along with Relu activation and 0.2 Dropout. These layers are then followed by a Dense Layer with 256 neurons and Relu followed by 4 output neurons and Softmax. Since Face\_Model2 achieves the best performance we use Face\_Model2 based architecture for the concatenated MoCap data architecture with 189 input feature length as Mocap\_Model1. The LSTM based models use Adadelta and Convolution and fully connected based models use Adam as optimizers.

\begin{table}[!t]
\renewcommand{\arraystretch}{1.3}
\caption{MoCap emotion detection models and accuracy}
\label{table_mocap}
\centering
\begin{tabular}{|c||c|}
\hline
Model & Accuracy\\
\hline
MoCap-head Head\_Model1 & 37.75\% \\
MoCap-head Head\_Model2 & 40.28\% \\
\hline
MoCap-hand Hand\_Model1 & 33.70\% \\
MoCap-hand Hand\_Model2 & 36.94\% \\
\hline
MoCap-face Face\_Model1 & 48.99\% \\
MoCap-face Face\_Model2 & 48.58\% \\
\hline
MoCap-combined Mocap\_Model1 & 51.11\% \\
\hline
\end{tabular}
\end{table}

\section{Results}
\label{sec:models}

\subsection{Combined Multi-Modal Emotion Detection}

\begin{figure}[!t]
\centering
\includegraphics[width=3.5in]{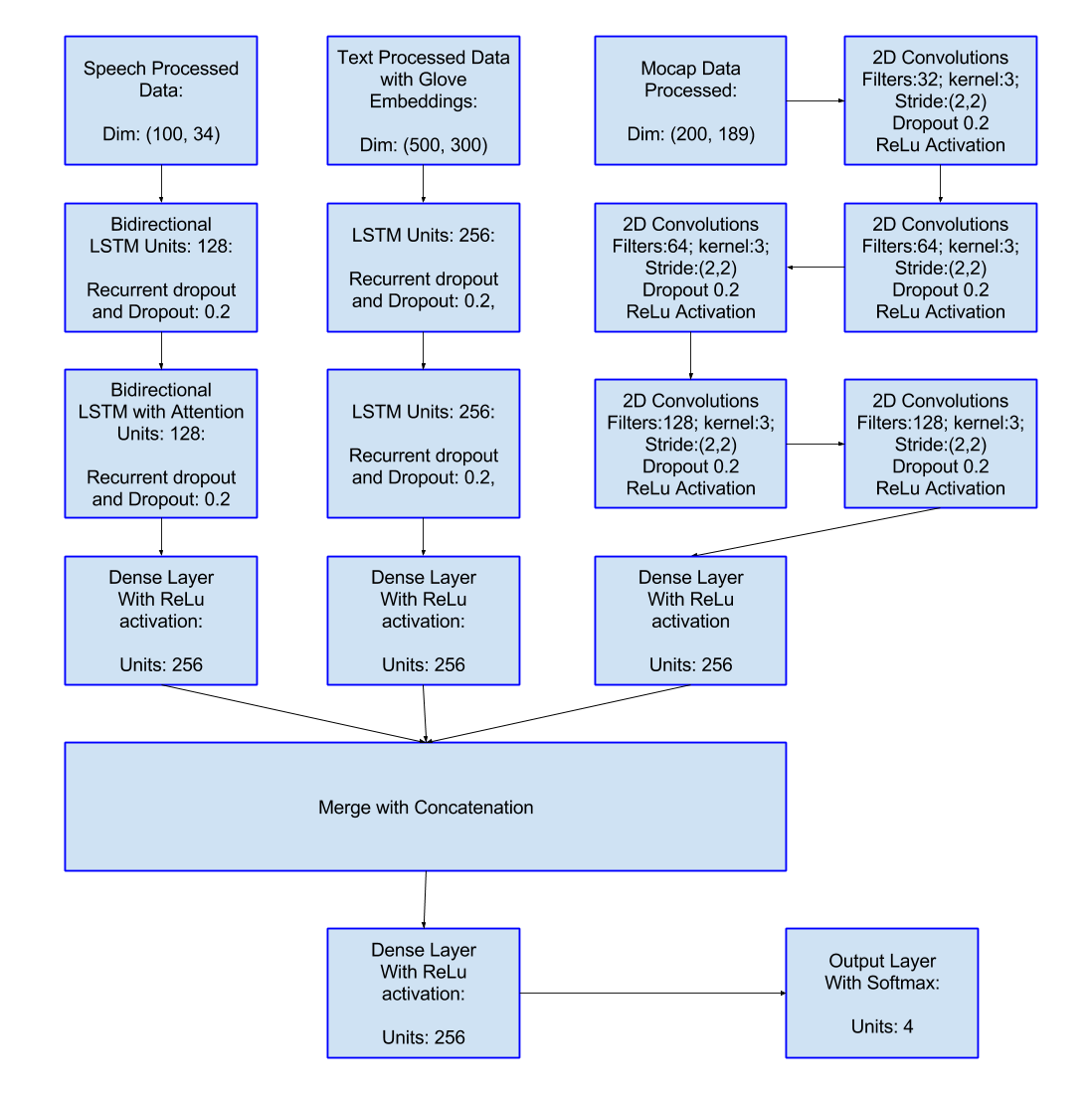}
\hfil
\caption{Final Combined Neural Network}
\label{com_sim}
\end{figure}

\begin{figure}[!t]
\centering
\includegraphics[width=3in]{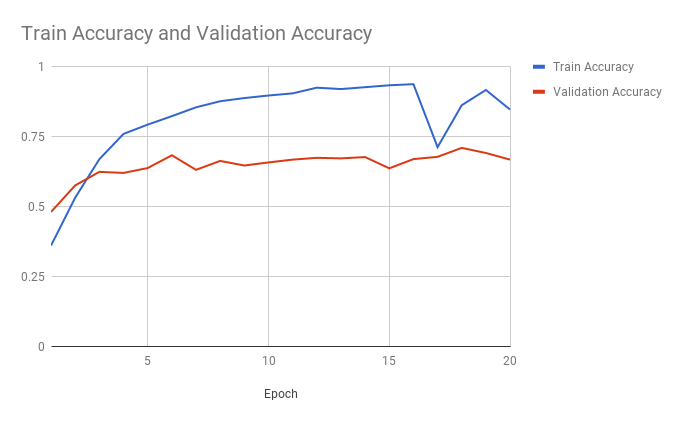}
\caption{Accuracy graph of our Final Model}
\label{acc_fig}
\end{figure}

\begin{table}[!t]
\renewcommand{\arraystretch}{1.3}
\caption{Multimodal emotion detection models and accuracy}
\label{table_mocap}
\centering
\begin{tabular}{|c||c|}
\hline
Model & Accuracy\\
\hline
Text + Speech + Mocap Combined & 71.04\% \\
Poria \cite{poria2018multimodal} & 71.59\% \\
\hline
\end{tabular}
\end{table}

For our final model we take the best individual models for each modality without their final softmax layers. The Text\_Model2 with stacked LSTMs and Glove word embeddings is chosen for text modality, Speech\_Model4 for the speech modality with 2 stacked bidirections LSTMs with Attention, and combined Mocap\_Model1 with stacked convolution layers. We then perform feature fusion by concatenating their final fully connected layers. We add another final fully-connected layer with 256 neurons followed by a softmax layer. This forms our combined final model.

We then perform hyperparamter optimization on this model. We choose the number of LSTM neurons in Speech\_Model4, LSTM neurons in Text\_Model2, neurons of the final fully connected layer of the combined model and net Dropout on all the models as hyperparameters. We use Random proposer to optimize training on the validation set. We then evaluate the best hyperparameter configuration on the test set, using train and validation set. Our performance matches the prior state of the art, however the comparison is not fair. \cite{poria2018multimodal} use 10 fold cross validation, while we use less training data in a 77\%-22\% split. 

\section{Conclusion}
\label{sec:concl}

In this paper we perform multimodal emotion recognition on IEMOCAP dataset using data from speech, text, and motions capture and identify best individual architectures for classification on each modality. We perform fusion only at the final layer which allows for a more robust and accurate emotion detection. Our approach has certain advantages. Firstly, since we only perform fusion at the final stage, lack of a modality would only require retraining the final fully connected layer. Also since we optimize individual modalities our combined model has a modular approach. This allows any individual model to be replaced by a better model, without affecting rest of the modalities in the combined model. Secondly, since we use motion captured data and 2D convolutions instead of video recordings and 3D convolutions, we have a faster training and inference time.

\bibliographystyle{IEEEbib.bst}
\bibliography{refs.bib}

\end{document}